# Design, Development and Testing of Underwater Vehicles: ITB Experience

**Mulyowidodo[1], Said D. Jenie[2], Agus Budiyono[3†], Sapto A. Nugroho[4]**

[1] *Automation & Robotics Laboratory, Department of Mechanical Engineering*
*Institut Teknologi Bandung*
*Jl. Ganesha 10, Bandung 40132, Indonesia*
*Tel: +62-22-2503775, Fax: +62-22-2503775, E-mail: muljo@bdg.centrin.net.id*

[2] *Department of Aeronautics & Astronautics*
*Institut Teknologi Bandung*
*Jl. Ganesha 10, Bandung 40132, Indonesia*
*Tel: +62-22-2504529, Fax: +62-22-2534164, E-mail: saiddj@bppt.go.id*

[3 †] *Corresponding author, Department of Aeronautics & Astronautics*
*Institut Teknologi Bandung*
*Jl. Ganesha 10, Bandung 40132, Indonesia*
*Tel: +62-22-2504529 ext. 206, Fax: +62-22-2534164, E-mail: agus.budiyono@ae.itb.ac.id*

[4] *Automation & Robotics Laboratory, Department of Mechanical Engineering*
*Institut Teknologi Bandung*
*Jl. Ganesha 10, Bandung 40132, Indonesia*
*Tel: +62-22-2503775, Fax: +62-22-2503775, E-mail: ssaptoadi@yahoo.com*

## Abstract

*The last decade has witnessed increasing worldwide interest in the research of underwater robotics with particular focus on the area of autonomous underwater vehicles (AUVs). The underwater robotics technology has enabled human to access the depth of the ocean to conduct environmental surveys, resources mapping as well as scientific and military missions. This capability is especially valuable for countries with major water or oceanic resources. As an archipelagic nation with more than 13,000 islands, Indonesia has one of the most abundant living and non-organic oceanic resources. The needs for the mapping, exploration, and environmental preservation of the vast marine resources are therefore imperative. The challenge of the deep water exploration has been the complex issues associated with hazardous and unstructured undersea and sea-bed environments. The paper reports the design, development and testing efforts of underwater vehicle that have been conducted at Institut Teknologi Bandung. Key technology areas have been identified and step-by-step development is presented in conjunction with the need to meet the challenge of underwater vehicle operation. A number of future research directions are also highlighted.*

## Introduction

Indonesia is a vast archipelago spanning three time zones with the area of 1,904,570 sq-km where about two-thirds are covered by the ocean. To enable the utilization of the untapped natural resources, a deep water surveying technology is a necessity. Underwater robotics technology can be deployed for the investigation of marine and environmental issues, the study of coastal dynamics and the protection of the oceanic resources from pollutants [1]. The use of unmanned underwater vehicle is essential to an improved understanding and characterization of physical and biological coastal dynamics in conjunction with the efficient and responsible utilization of natural resources for human welfare. The underwater vehicle platform with a capability of resolved continuous measurements will be instrumental in mapping the oceanic resources for well-planned resources exploitation such as fisheries, mining, oil and gas drilling activities taking into account the need to maintain a long-term marine ecological balance. Unlike other fixed platforms, autonomous underwater vehicles are particularly of interest due to their ability to provide continuous spatial and temporal observations. Also, in contrast to satellite and airborne sensors which are limited to surface observations, AUVs have the capability of characterizing dynamics in the entire water column [2].

## Approach and Methods

Most commercial unmanned underwater vehicles are tethered and remotely operated, referred to as remotely piloted vehicles (ROVs)[1]. The demand for the autonomous underwater vehicles has been growing because of the fact that other platforms such as the manned submersibles, towed instrument sleds and ROVs are limited to a few applications due to the associated high operation cost and risks. The technology of AUVs dated back to the early 1960s at the Applied Physics Laboratory at the University of Washington with the Self-Propelled Underwater Research Vessel (SPURV) [2]. Since then advances have been made by numerous worldwide research centers and universities. Ref.[1] gives an extensive review on the design and control technology for the autonomous underwater vehicles where key subsystems were identified, recent subsystem developments were surveyed and current state-of-the-art were summarized. Based on the study, there are more than 46 AUV models in 2000, the year when the survey was conducted. In the 1990s, a variety of AUVs were designed for different purposes such as underwater test-bed, water column inspector, mine countermeasures, precision control platform, search and mapping, bottom/sub-bottom survey, pipeline/cable inspection, environmental monitoring, military mission, and undersea shuttle. Along the design evolution, key technology areas have been manifested in the guidance[15], navigation[6,9,10,17], control[1,7,11,14,16], communication, power[5], pressure halls/fairings, and mechanical manipulator systems.

Ref.[2] specifically surveyed the trends in bio-robotic autonomous undersea vehicle. The significant advances in three disciplines, namely the biology-inspired high-lift unsteady hydrodynamics, artificial muscle technology and neuroscience-based control, are discussed in an effort to integrate them into viable products. The understanding of the mechanisms of delayed stall, molecular design of artificial muscles and the neural approaches to the actuation of control surfaces is reviewed in the context of devices based on the pectoral fins of fish, while remaining focused on their integrated implementation in biorobotic Autonomous Undersea Vehicles [2].

The ongoing research activities are aiming at enhancing the autonomy of the underwater vehicle including better design of communication, higher power density and more reliable navigation and control for deep water operation. The existing primary methods for AUVs navigation are [6]: (1) dead-reckoning and inertial navigation systems, (2) acoustic navigation, and (3) geophysical navigation techniques. The use of dead-reckoning and inertial navigation system (INS) has been inhibited by the high cost and power consumption especially for small AUVs. Lower grade INS on the other hand poses a problem of error drift as the vehicle travels further distance. An integration of INS with other sources of error-bounding navigation such as Doppler velocity sonar (DVS) or GPS through Kalman filtering is desirable and has been proven to be a viable solution. Unlike the tethered ROVs that are powered by the mother ship, the AUVs depend on the power traditionally provided by lead-acid type battery. Due to higher energy density, ten to twenty-fold as high, fuel-cell and fuel-cell-like devices have been attracted more attention in the area of AUV power.

Due to the complex nature of the underwater vehicle dynamics, oceanic disturbance and uncertainty pertaining to changes in center of gravity and buoyancy, AUVs demand control system that has a self-tuning ability. Numerous approaches to control strategy have been developed to address the need including sliding mode control, neural network, fuzzy control and adaptive control. Ref[14] reports comparative study of various control techniques for underwater vehicle used in the 1990s. The PID control was reported to be successful in the provision that the vehicle is operated under a constant speed thus justifying the linear approach. The comparison between $H\infty$ controller and classical approach reveals that the two achieve similar performance making the former technique unattractive due to significantly more complex design procedure. The study concludes that the more obvious control candidates for initial investigation are classical controllers, fuzzy logic and sliding mode.

In this paper, we present a bird eye view of the autonomous underwater vehicles developed at Institut Teknologi Bandung in the research span of 2001 to present. The principle of operation including the technical specification of the subsystem is conveyed. The navigation and control strategy is also described as illustration. A field test demonstrates the viability of the vehicle to be used as underwater robotics for a variety of missions.

## Results

### Period of 2001-2002

The first prototype of the underwater vehicle is designed as a test-bed with operating depth of up to 10 m with a cruising speed of 3 knots. The vehicle is powered by 24 VDC power supply and 12V Lead Acid battery. The 3D drawing is given by Fig. 1. The technical specification is presented in Table 1.

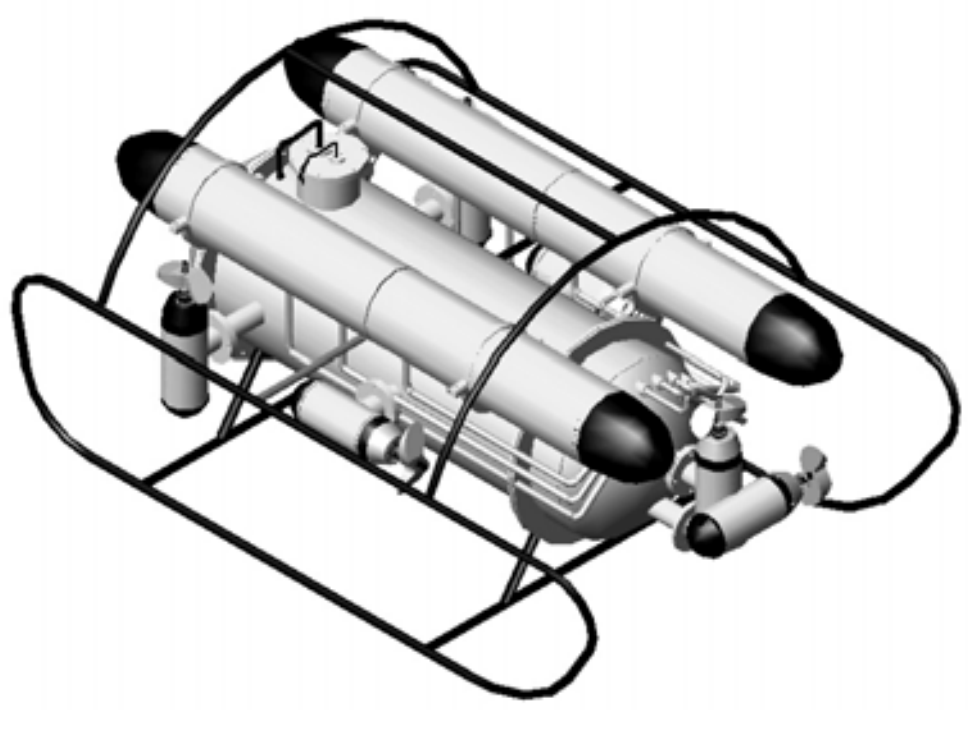

Fig. 1 ITB-BPPT AUV 2001-2002 design

The sensor suit contains gyro, MLDA, depth sensor, camera and leakage detector. The position information, leak detection and power distribution are sent to fault manager which eventually transmit the signal to maneuvering control unit and communication unit for display to the remote operator. The maneuvering unit receives information from mission plan through the mission executor. The maneuver can be achieved using the buoyancy control by means of control valve and using the propulsion control by means of motor driver controller. The remote operator is equipped with user interface consisting joystick, keyboard and display monitor all connected through cabling system using Belden utp cable for data and Eterna 2.5mm 4 wires for power. To help the vehicle's vision, three 50 W Halogen lamp were used. The control system is realized using 1 PC with Pentium III processor and 192 RAM running Borland C++ and 5 microcontrollers from Atmel running CVAVR with C compiler. The communication is conducted through multipoint RS 485 protocol. The schematic of the control system for the first prototype underwater vehicle is illustrated in Fig. 2.

**Period of 2003-2004**

The second prototype features a more advanced underwater vehicle design with the operating depth of up to 300m and the speed of 4 knots. Fig. 3 shows the drawings of the vehicle dimensioned at 1200 mm (L) x 800 mm (W) x 800 mm (H) and weighed 150 kg. The orientation is obtained through triad accelerometers, gyros and magnetometers. While the depth and leakage is measured and detected respectively by the same transducer as those of the first prototype vehicle. The design is equipped with hydraulically actuated 4 axis manipulator with the maximum payload of 10 kg.

Due to space limitation the control system diagram is not shown.

Table 1 – Technical specification of ITB-BPPT 1st prototype underwater

| **Specs** | **Value** |
|---|---|
| Speed | 3 knots |
| Weight in air | 43.4 kg |
| Dimension | 1,080x730x560 mm |
| Operating depth | 10 m |
| Thruster | 6 Brush DC motor 3A/24 V, Preamp and Amplifier motor driver |
| Orientation sensor | 2 MLDA(Mercury Level Detector Array) for roll and pitch angle 1 SMM (*Silicone Micro Machine*) Gyro from Futaba (GY-240) with AVCS(*Angular vector control system*) for yaw angle |
| Depth sensor | Omron E8CA-R8 (Pressure Sensor) |
| Leakage sensor | 8 position TFCLD (Thin Film Contact Leak Detector) |
| Camera | 12mm B/W digital camera |

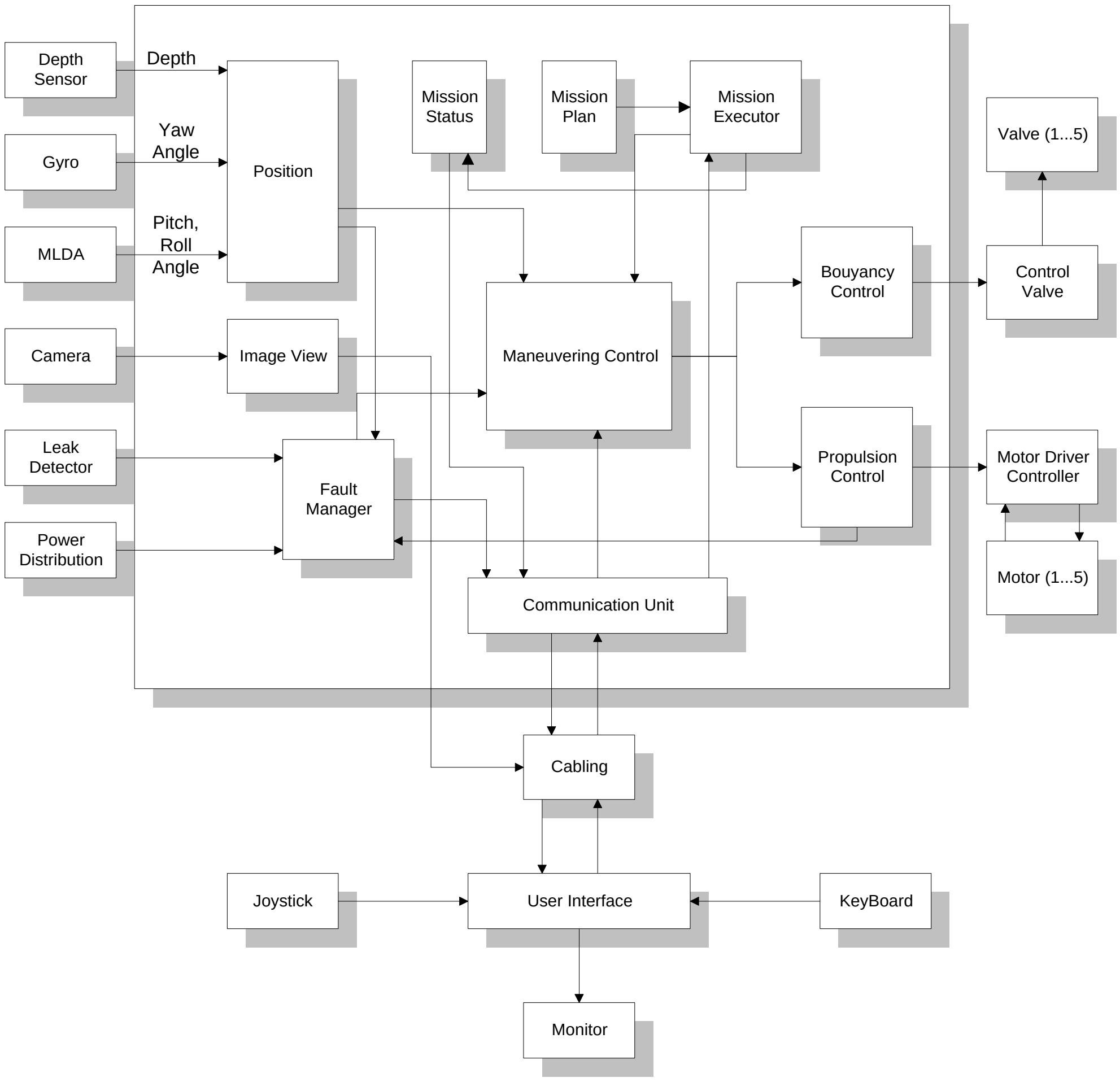

Fig. 2 The control system for ITB-BPPT AUV 2001-2002 design

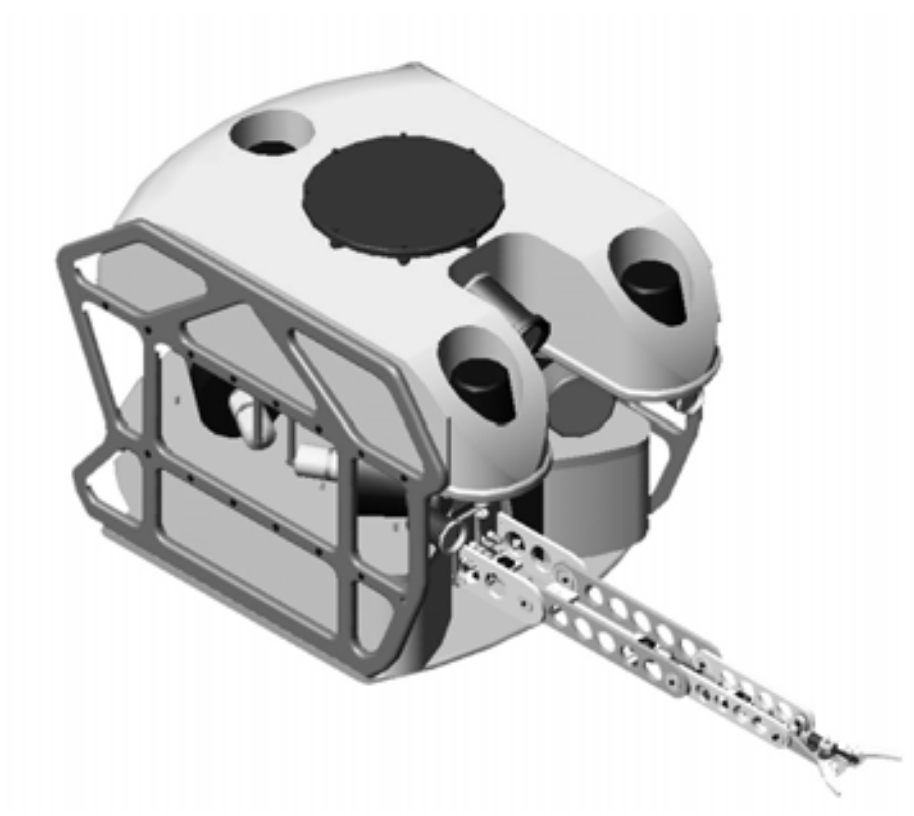

Fig. 3 ITB-BPPT AUV 2003-2004 design

### Period of 2005-now

The third prototype is biologically-inspired design characterized by squid-like structure for a better hydrodynamic property shown by Fig. 4. The more refined prototype is supported by more elaborate sensor arrays which consist of triad accelerometers, gyros and magnetometers for position and orientation, speed log sensor for velocity measurement, single beam altimeter, depth sensor, GPS, Doppler Velocity Log (DVL), and USBL tracking system. The vehicle weighs 300 kg with the dimension of 4400mm x 750 mm x 950 mm. The operating depth is up to 100 m within the working range of 6 nautical miles and the cruising speed of 6 knots. The thrust is provided by 3 vectoring position (@ 1 Hp) Technadyne Brushless DC thruster. The control architecture is given by Fig. 5.

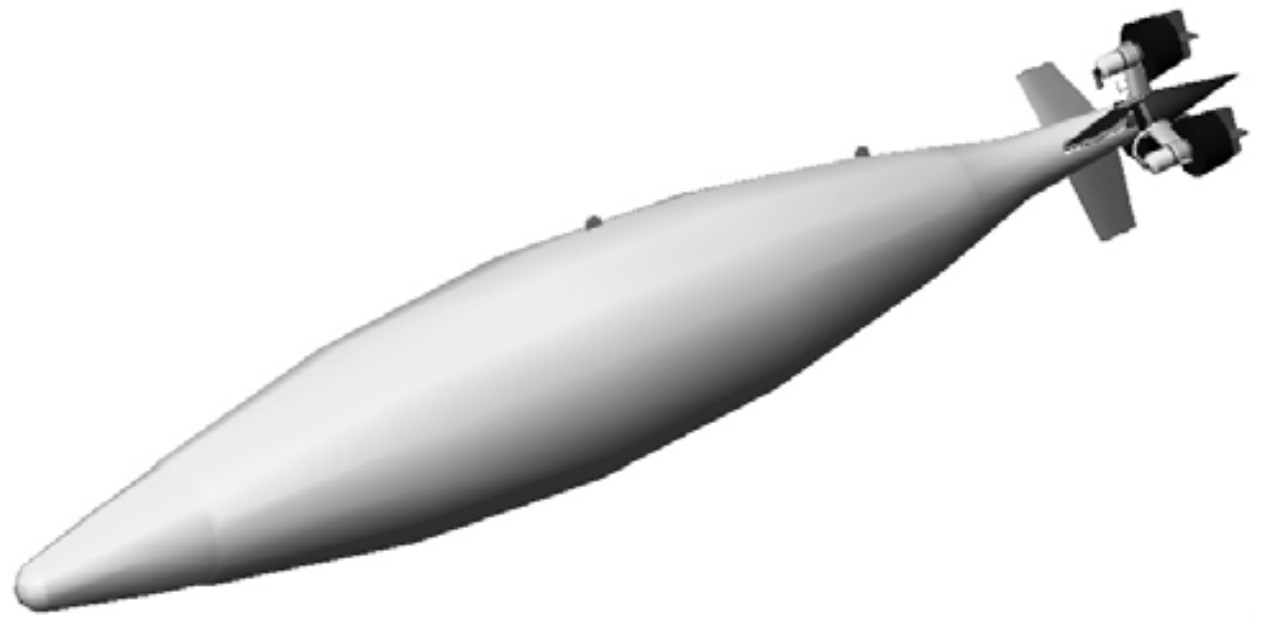

Fig. 4 ITB-BPPT AUV 2005 design named Sotong (squid)

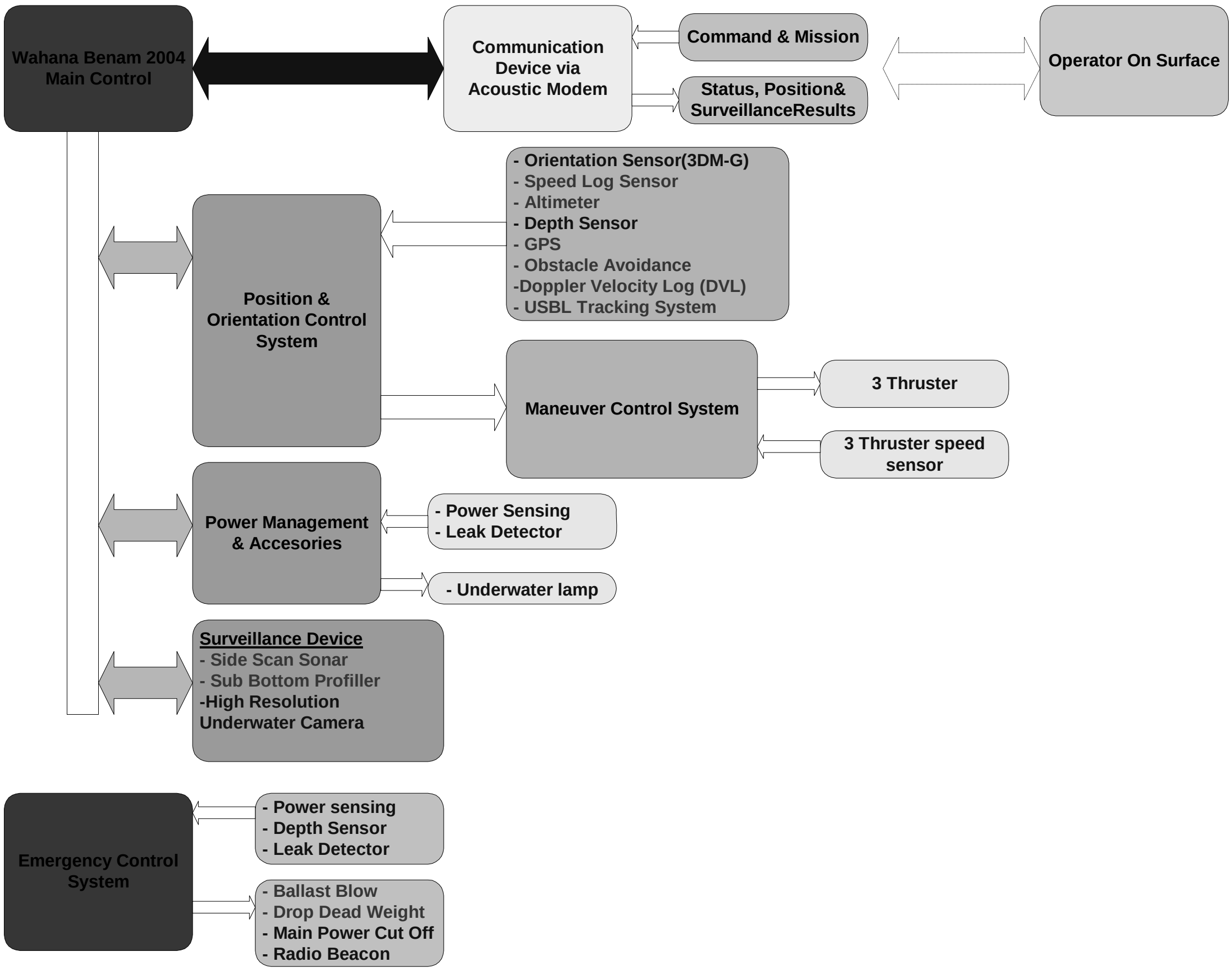

Fig. 5 The control architecture of ITB-BPPT AUV 2005 design

## Discussion

### Non-autonomous operation

A series of testing were conducted to assess the performance of the underwater vehicle designs. The testing of the first prototype was conducted in the shallow water environment as the operating depth is only up to 10 meter (Fig. 6). Even though the test was performed in the limited depth, it serves to primary purpose of validating the basic control and communication system.

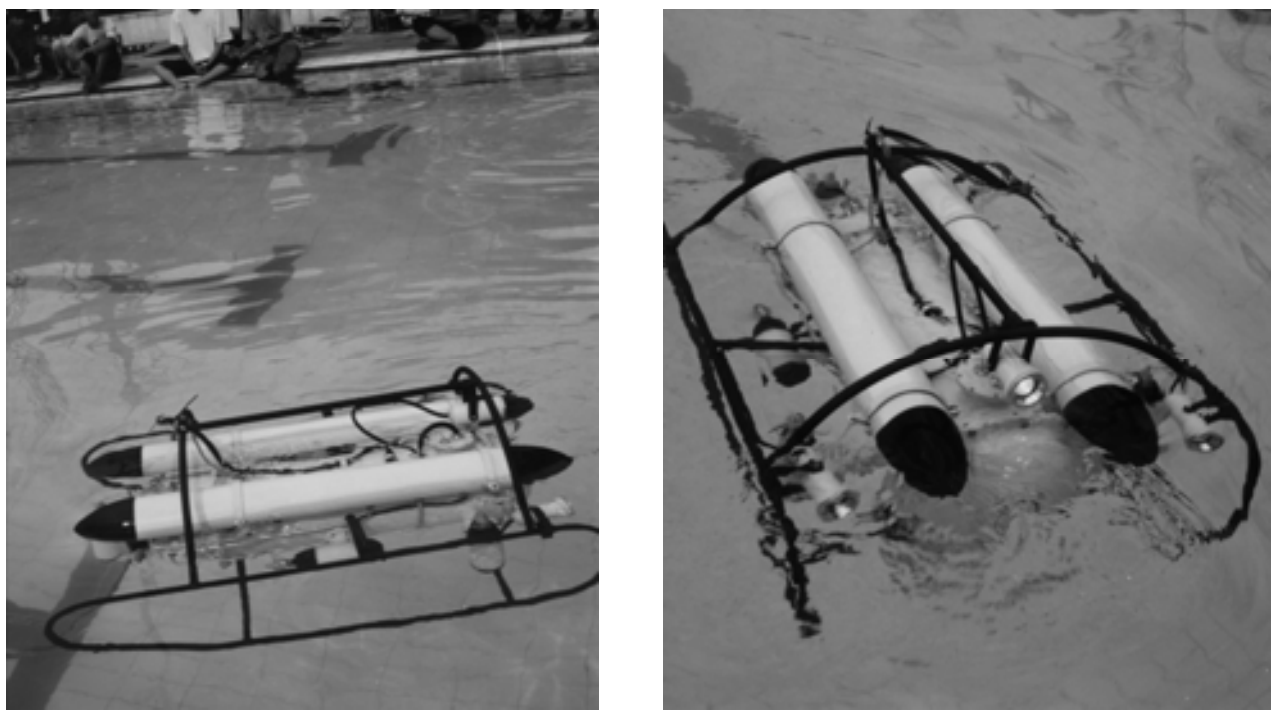

Fig.6 Initial test of first prototype ITB-BPPT AUV

The lessons learned from the testing of the first prototype have been used to carry out the field testing for the second prototype conducted in the ocean. Fig. 7 shows the activities of the field testing of the second prototype vehicle.

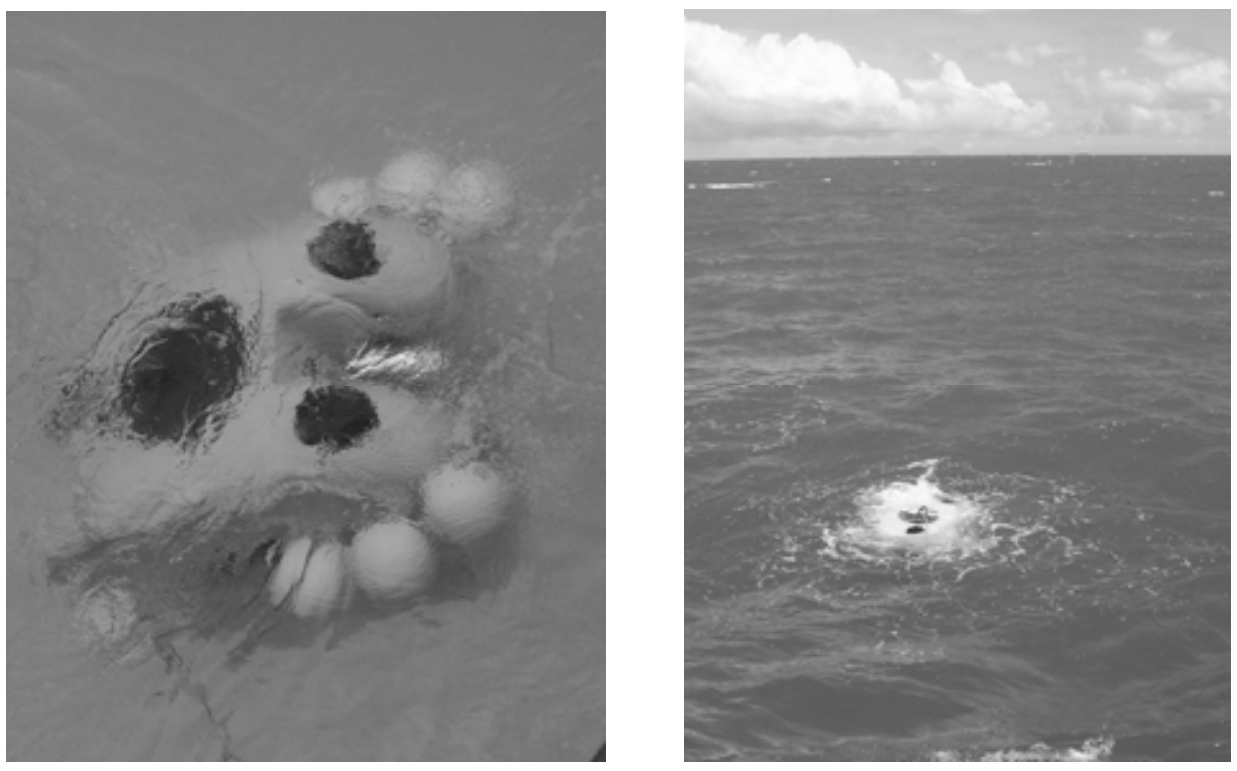

Fig. 7 Ocean test of the second prototype ITB-BPPT AUV

### Autonomous operation

An autonomous control system has been design for the third ITB AUV prototype, Sotong. The autonomous maneuvering tests have been conducted in the indoor towing tank at LHI-BPPT (Hydrodynamics Laboratory) in Surabaya. Beyond the standard yawing, circle and rectangular maneuvers, the autonomous maneuvers also include tracking of a growing and contracting reversed spiral (Dieudonne's maneuver) trajectory. The attitude angle time history during the autonomous Dieudonne's maneuver is presented in Fig. 8 as illustration. The good tracking performance for difficult predefined trajectory shows promising preliminary results for further development of the autonomous navigation and control of the ITB-BPPT AUV.

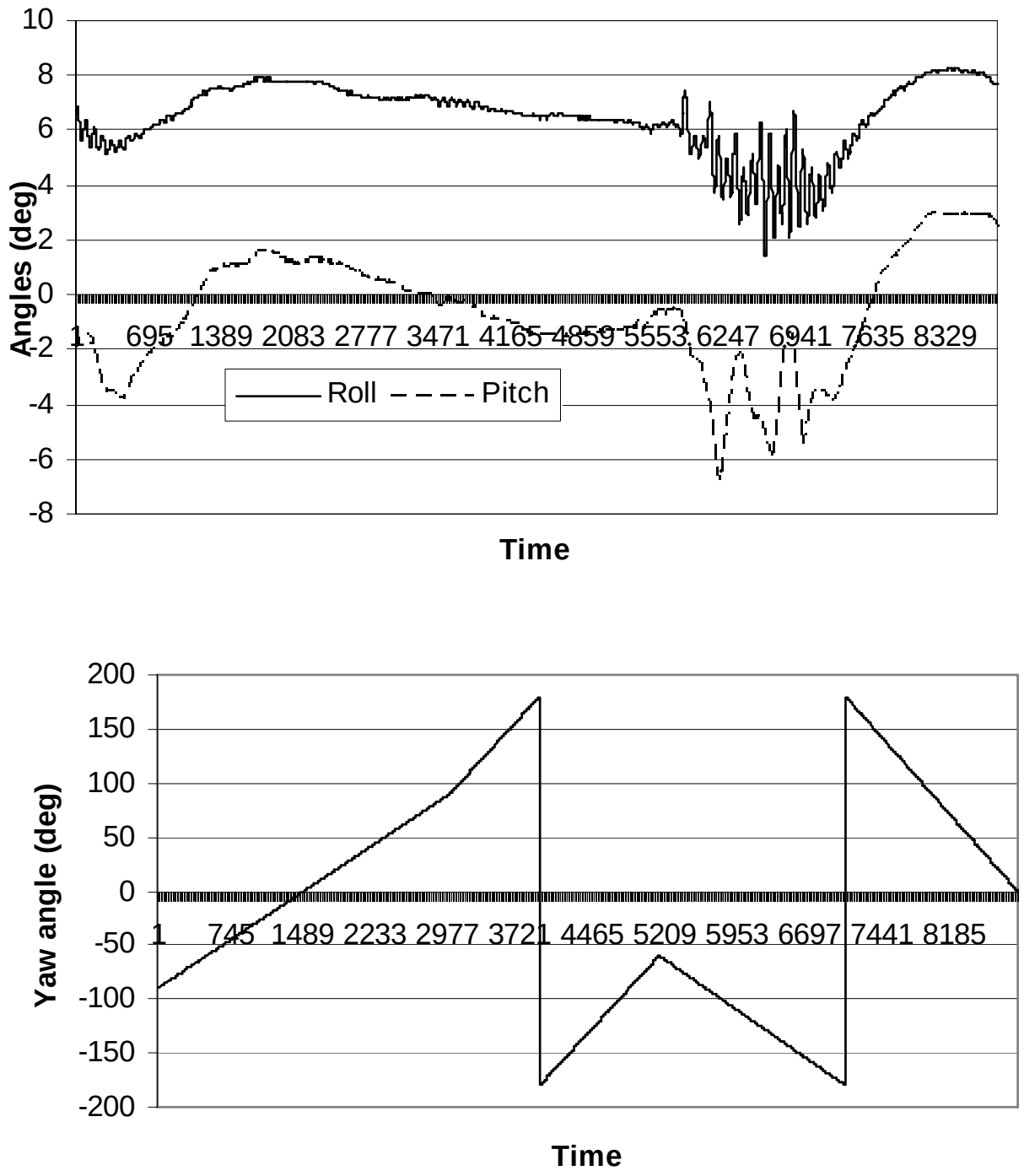

Fig.8 Time history of attitude angles during autonomous Dieudonne's maneuver

## Concluding Remarks

The paper presents the overview of the underwater vehicle research program at Institut Teknologi Bandung. A number of progress in the design, development and testing of the AUVs are highlighted. The technical specification and control architecture of the prototypes are presented as illustration. Using the available vehicles as research platform, further work will be carried out in the area of control and navigation techniques development including the application of sensor fusion technology for a better navigation, investigation of various control techniques and the potential use of biologically-inspired robotic mechanisms.

One anticipated challenge in the AUV navigation is the uncertainty in vehicle position estimation due to the fact that the position information is updated only every few seconds. It will be critical both for target approach and particularly collision avoidance task. In light of this, a technique to integrate sensors to yield high performance motion estimation will be of interest. Further effort will include the integration of multi-array sonar through Extended Kalman Filtering algorithm. In addition, the integration of other sensors such as current meter, Doppler sonar and monocular video system is envisioned as part of autonomous collision avoidance system development. Future efforts will be ultimately aimed at the integration of the learned technologies for manufacturing a viable product to meet the demand of efficient exploration and utilization of Indonesia's vast oceanic resources.

## Acknowledgments

The work was supported by the Research Grant made available by the Technology Assessment and Application Agency (BPPT) of Indonesia. The authors would like to thank the technical team involved through the design process, manufacturing and testing of the vehicles.